\documentclass[conference]{IEEEtran}
\IEEEoverridecommandlockouts
\usepackage{CJKutf8}
\usepackage{pinyin}
\usepackage{cite}
\usepackage{amsmath,amssymb,amsfonts}
\usepackage{algorithmic}
\usepackage{algorithm}
\usepackage{graphicx}
\usepackage{textcomp}
\usepackage{xcolor}
\usepackage{bm}
\usepackage{multirow}

\def\BibTeX{{\rm B\kern-.05em{\sc i\kern-.025em b}\kern-.08em
    T\kern-.1667em\lower.7ex\hbox{E}\kern-.125emX}}
\begin{document}

\title{Fine-tuning BERT for Joint Entity and Relation Extraction in Chinese Medical Text}
\author{\IEEEauthorblockN{Kui Xue$^1$, Yangming Zhou$^{1,*}$, Zhiyuan Ma$^{1}$, Tong Ruan$^{1}$, Huanhuan Zhang$^{1,*}$ and Ping He$^{2}$}
\IEEEauthorblockA{
$^1$School of Information Science and Engineering, East China University of Science and Technology, Shanghai 200237, China \\\
$^2$Shanghai Hospital Development Center, Shanghai 200041, China\\
$^*$Corresponding authors\\
Emails: \{ymzhou,hzhang\}@ecust.edu.cn\\}
}

\maketitle

\begin{abstract}

Entity and relation extraction is the necessary step in structuring medical text. However, the feature extraction ability of the bidirectional long short term memory network in the existing model does not achieve the best effect. At the same time, the language model has achieved excellent results in more and more natural language processing tasks. In this paper, we present a focused attention model for the joint entity and relation extraction task. Our model integrates well-known BERT language model into joint learning through dynamic range attention mechanism, thus improving the feature representation ability of shared parameter layer. Experimental results on coronary angiography texts collected from Shuguang Hospital show that the F$_1$-scores of named entity recognition and relation classification tasks reach 96.89\% and 88.51\%, which outperform state-of-the-art methods by 1.65\% and 1.22\%, respectively.

\end{abstract}


\section{Introduction}

With the widespread of electronic health records (EHRs) in recent years, a large number of EHRs can be integrated and shared in different medical environments, which further support the clinical decision making and government health policy formulation\cite{gunter2005emergence}. However, most of the information in current medical records is stored in natural language texts, which makes data mining algorithms unable to process these data directly. To extract relational entity triples from the text, researchers generally use entity and relation extraction algorithm, and rely on the central word to convert the triples into key-value pairs, which can be processed by conventional data mining algorithms directly. 

\begin{figure}
\begin{center}\includegraphics[width=0.20\textwidth]{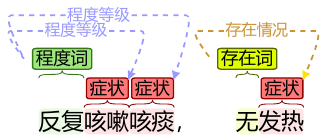}
\caption{An illustrative example of entity and relation extraction in the text of EHRs.}
\label{fig:re-example}
\end{center}
\end{figure}

To solve the task of entity and relation extraction, researchers usually follow pipeline processing and split the task into two sub-tasks, namely named entity recognition (NER) and relation classification (RC), respectively. However, this pipeline method usually fails to capture joint features between entity and relationship types. \begin{CJK}{UTF8}{gkai}For example, for a valid relation “存在情况(presence)” in Fig.~\ref{fig:re-example}, the types of its two relational entities must be “疾病(disease)”, “症状(symptom)” or “存在词(presence word)”.\end{CJK} To capture these joint features, a large number of joint learning models have been proposed \cite{zheng-etal-2017-joint,sun2019joint}, among which bidirectional long short term memory (Bi-LSTM) \cite{zheng2017joint} are commonly used as the shared parameter layer. However, compared with the language models that benefit from abundant knowledge from pre-training and strong feature extraction capability, Bi-LSTM model has relatively lower generalization performance. To improve the performance, one of the solutions is to incorporate language model into joint learning as a shared parameter layer. However, existing models only introduce language models into the NER or RC task separately \cite{dogan2019fine,alt2019improving}, leading the joint features between entity and relationship types unable to be captured.

%Therefore, the joint features between entity and relationship types still can not be captured.

Given the aforementioned challenges and current researches, we propose a focused attention model based on widely known BERT language model \cite{devlin-etal-2019-bert} to jointly tackle NER and RC tasks. Specifically, through the dynamic range attention mechanism, we construct task-specific MASK matrix to control the attention range of the last $K$ layers in BERT language model, leading to the model focusing on the words of the task. This process helps obtain the corresponding task-specific context-dependent representations. In this way, the modified BERT language model can be used as the shared parameter layer in joint learning of NER and RC task. We call the modified BERT language model as shared task representation encoder (STR-encoder) in the following paper. The main contributions of our work are summarized as follows:
\begin{itemize}
    \item We propose a focused attention model to jointly learn NER and RC task. The model integrates BERT language model as a shared parameter layer to achieve better generalization performance.
    \item In the proposed model, we incorporate a novel structure, called STR-encoder, which changes the attention range of the last $K$ layers in BERT language model to obtain task-specific context-dependent representations. It can make full use of the original structure of BERT to produce the vector of the task, and directly use the prior knowledge contained in the pre-trained language model.
    \item For RC task, we design two different MASK matrices to extract the required feature representation of RC task. The performances corresponding to the matrices are analyzed and compared in the experiment.
\end{itemize}

\section{Related Work}
\label{section:related-work}

Entity and relation extraction is to extract relational entity triplets. There are two kinds of approaches for the task, namely Pipeline and joint learning methods. The former decomposes the task into two subtasks, namely named entity recognition (NER) and relation classification (RC), while the latter attempts to solve the two tasks simultaneously.

\subsection{Named Entity Recognition}

In medical domain, we use NER to recognize disease, symptom, etc. In general, NER is formulated as a sequence tagging task using BIEOS (Begin, Inside, End, Outside, Single) \cite{Krishnan2005NamedER} tagging strategy. Conventional methods in medical domain can be divided into two categories, i.e., statistical and neural network methods. The former are generally based on conditional random fields (CRF) \cite{skeppstedt2014automatic} which relies on hand-crafted features and external knowledges to improve the accuracy. Neural network methods typically use neural network to calculate the features without tedious feature engineering, e.g., bidirectional long short term memory neural network \cite{wang2019incorporating}. However, none of the above methods can make use of a large amount of unsupervised corpora, resulting in limited generalization performance.

\subsection{Relation Classification}

RC is closely related to NER task, which classifies the relationship between the entities identified in the text. The task is typically formulated into a classification problem that takes a piece of text and two entities in this text as inputs, and possible relation between the entities as output. The existing methods of RC can be roughly divided into two categories, i.e., traditional methods and neural network approaches. The former are based on feature-based \cite{rink2010utd} or kernel-based \cite{zelenko2003kernel} approaches. These models usually spend a lot of time on feature engineering. Neural network methods can extract the relation features without complicated feature engineering. e.g., recurrent capsule network \cite{wang2018automatic} and domain invariant convolutional neural network \cite{sahu2016relation}. However, These methods cannot utilize joint features between entity and relation, resulting in lower generalization performance when compared with joint learning methods.

\subsection{Joint Entity and Relation Extraction}

%Joint entity and relation extraction tasks solve NER and RC simultaneously. 
Compared with pipeline methods, joint learning approaches are able to capture the joint features between entities and relations \cite{li2014incremental}.

State-of-the-art joint learning methods can be divided into two categories, i.e., joint tagging methods and parameter sharing methods. Joint tagging methods transform NER and RC tasks into sequence tagging tasks through a specially designed tagging scheme, e.g., a novel tagging scheme proposed by Zheng et al. \cite{zheng-etal-2017-joint}. Parameter sharing methods share the feature extraction layer in the models of NER and RC. Compared to joint tagging methods, parameter sharing methods are able to effectively process multi-map problem. The most commonly shared parameter layer in medical domain is the Bi-LSTM network \cite{li2017neural}. However, compared with language model, the feature extraction ability of Bi-LSTM is relatively weaker, and the model cannot obtain pre-training knowledge through a large amount of unsupervised corpora, which further reduces the robustness of extracted features.

\section{Proposed Method}
\label{section:proposed-method}

In this section, we first introduce classic BERT language model and the dynamic range attention mechanism. Then, we present a focused attention model for joint entity and relation extraction.

\subsection{BERT Language Model}

BERT \cite{devlin-etal-2019-bert} is a language model that utilizes bidirectional attention mechanism and large-scale unsupervised corpora to obtain effective context-sensitive representations of each word in a sentence. Owing to its effective structure and a rich supply of large-scale corpora, BERT has achieved state-of-the-art results on various natural language processing (NLP) tasks. The basic structure of BERT includes self attention encoder (SA-encoder) and downstream task layer. SA-encoder obtains the corresponding context-dependent representation using the sequence $S$ and the MASK matrix:
\begin{equation}
    H_N = SA\verb|-|encoder(S,MASK)
\end{equation}

The downstream task layer differs from task to task. In this work, we focus on NER and RC, which are further detailed in Section \ref{section:construction_downstream_ner} and \ref{section:construction_downstream_rc}, respectively.

\subsection{Dynamic Range Attention Mechanism}
\label{section:dynamic-attention-range}
In BERT, MASK matrix is originally used to mask the padding portion of the text. However, we observe that, with the help of a specific MASK matrix, we can directly control the attention range of each word, thus obtaining specific context-sensitive representations.

Note that, when calculating the attention in BERT, the parameter matrix MASK$\in {\{0,1\}}^{T\times T}$, where $T$ is the length of the sequence. If MASK$_{i,j} = 0$, then we have (MASK$_{i,j}-1$)$\times \infty = -\infty$ and the Eq. \eqref{Similar,Mask=0}, which indicates that the $i$-th word ignores the $j$-th word when calculating attention.
\begin{equation}
\begin{aligned}
\label{Similar,Mask=0}
   & Similar(i,j) \\
    &= Softmax[{\frac{QK^T}{\sqrt{d_k}}}_{i,j}+(MASK_{i,j}-1)\times\infty]\\
    &= Softmax(-\infty)
\end{aligned}
\end{equation}

When MASK$_{i,j} = 1$, we have $(MASK_{i,j}-1)\times \infty = 0$ and the Eq. \eqref{Similar,Mask=1}, which means the $i$-th word considers the $j$-th word when calculating attention.
\begin{equation}
\begin{aligned}
\label{Similar,Mask=1}
    & Similar(i,j) \\
    &= Softmax[{\frac{QK^T}{\sqrt{d_k}}}_{i,j}+(MASK_{i,j}-1)\times\infty]\\
    &=Softmax({\frac{QK^T}{\sqrt{d_k}}}_{i,j})
\end{aligned}
\end{equation}

\subsection{Focused Attention Model}
\label{section:focused-attention-model}
\begin{figure}
\begin{center}
\includegraphics[width=0.35\textwidth]{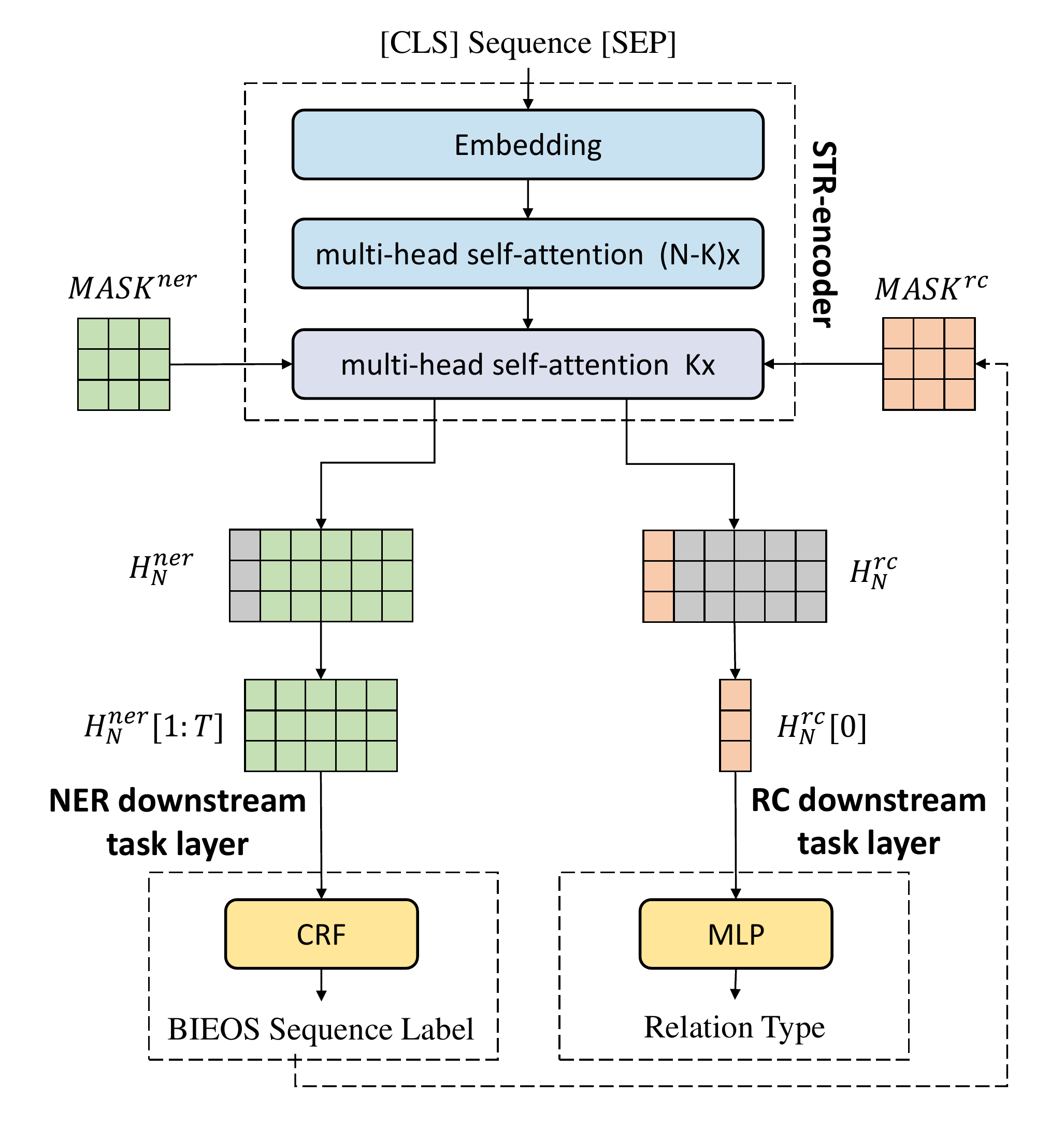}
\caption{The architecture of our proposed model.}
\label{fig:overview}
\end{center}
\end{figure}

The architecture of the proposed model is demonstrated in the Fig. \ref{fig:overview}. The focused attention model is essentially a joint learning model of NER and RC based on shared parameter approach. It contains layers of shared parameter, NER downstream task and RC downstream task.

The shared parameter layer, called shared task representation encoder (STR-encoder), is improved from BERT through dynamic range attention mechanism. It contains an embedded layer and $N$ multi-head self-attention layers which are divided into two blocks. The first $N-K$ layers are only responsible for capturing the context information, and the context-dependent representations of words are expressed as $H_{N-K}$. According to characteristics of NER and RC, the remaining K layers use the MASK$^{task}$ matrix setting by the dynamic range attention mechanism to focus the attention on the words. In this manner, we can obtain task-specific representations $H_N^{task}$ and then pass them to corresponding downstream task layer. In addition, the segmentation point $K$ is a hyperparameter, which is discussed in Section \ref{section:hyperparameter-analysis}.

Given a sequence, we add a $[CLS]$ token in front of the sequence and a $[SEP]$ token at the end of the sequence as BERT does. After the Embedding layer, the initial vector of each word in the sequence $S$ is represented as $H_0$, which is same as BERT. Then we input $H_0$ to the first $N-K$ multi-head self-attention layers.
In theses layers, attention of a single word is evenly distributed on all the words in the sentence to capture the context information. Given the output (${H}_{m-1}$) from the $(m-1)$-th layer, the output of current layer is calculated as:
\begin{align}
    &H_{m}'=LN[H_{m-1}+{MHSA}(H_{m-1},{MASK}^{all})] \\
    &H_m=LN[H_{m}'+PosFF(H_{m}')]
\end{align}
where $MHSA$, $PosFF$ and $LN$ represent multi-head self attention, feed forward and layer normalization \cite{vaswani2017attention} and MASK$^{all}\in {\{1\}}^{T\times T}$ indicates each word calculates attention with all the other words of the sequence.

The remaining $K$ layers focus on words of downstream task by task-specific matrix MASK$^{task}$ based on dynamic range attention mechanism.
Given the output ($H_{m-1}^{task}$) of previous $(m-1)$-th layer, the current output ($H_m^{task}$) is calculated as:
\begin{align}
    &H_m^{'task} = LN[H_{m-1}^{task}+{MHSA}(H_{m-1}^{task},{MASK}^{task})] \\
    &H_m^{task} = LN[H_m^{'task} +PosFF(H_m^{'task})]
\end{align}
where $H_{N-K}^{task} =H_{N-K}$ and $task\in \{ner,rc\}$.

As for STR-encoder, we only input different MASK$^{task}$ matrices, which calculate various representations of words required by different downstream task ($H_N^{task}$) with the same parameters:
\begin{equation}
\begin{aligned}
    H_N^{task} = STR\verb|-|encoder(S,MASK^{task},MASK^{all})
\end{aligned}
\end{equation}
The structure has two advantages:
\begin{itemize}
    \item It obtains the representation vector of the task through the strong feature extraction ability of BERT. Compared with the complex representation conversion layer, the structure is easier to optimize.
    \item It does not significantly adjust the structure of the BERT language model, so the structure can directly use the prior knowledge contained in the parameters of pre-trained language model.
\end{itemize}

Subsequently, we will introduce the construction of MASK$^{task}$ and downstream task layer of NER and RC in blocks.

\subsubsection{The Construction of MASK$^{ner}$}

In NER, the model needs to output the corresponding BIEOS tag of each word in the sequence. In order to improve the accuracy, the appropriate attention weight should be learned through parameter optimization rather than limiting the attention range of each word. Therefore, according to the dynamic range attention mechanism, the value of the MASK$^{ner}$ matrix should be set to MASK$_{ner}\in {\{1\}}^{T\times T}$, indicating that each word can calculate attention with any other words in the sequence.

\subsubsection{The Construction of NER Downstream Task Layer}
\label{section:construction_downstream_ner}

In NER, the downstream task layer needs to convert the representation vector of each word in the output of STR-encoder into the probability distribution of the corresponding BIEOS tag. Compared with the single-layer neural network, CRF model can capture the link relation between two tags \cite{Huang2015Bidirectional}. As a result, we perform CRF layer to get the probability distribution of tags. Specifically, the representation vectors of all the words except $[CLS]$ token in the output of STR-encoder are sent to the CRF layer. Firstly, CRF layer calculates the emission probabilities by linearly transforming these vectors. Afterwards, layer ranks the sequence of tags by means of emission and transition probabilities. Finally, the probability distribution of sequence of tags is obtained by softmax function:
\begin{equation}
\begin{aligned}
    &H_p^{ner} = H_N^{ner}[1:T]\times W_{ner}^{*}+b_{ner}
\end{aligned}
\end{equation}
\begin{equation}
\begin{aligned}
    &Score(L|H_p^{ner}) = \sum_{t=1}^{T}(A_{L_{t-1},L_t}+{H_p^{ner}}_{t,L_t})
\end{aligned}
\end{equation}
\begin{equation}
\begin{small}
\begin{aligned}
    &p_{ner}(L|S,MASK^{ner},MASK^{all})\!=\!\frac{e^{Score(L|H_p^{ner})}}{\sum_{J}e^{Score(J|H_p^{ner})}}
\end{aligned}
\end{small}
\end{equation}

The loss function of NER is shown as Eq. \eqref{equation:loss_ner}, and our training goal is to minimize $L_{ner}$, where $L'$ indicates the real tag sequence.
\begin{equation}
\label{equation:loss_ner}
\begin{aligned}
    &L_{ner} = -Log[p_{ner}(L'|S,MASK^{ner},MASK^{all})]
\end{aligned}
\end{equation}

\subsubsection{The Construction of MASK$^{rc}$}
\label{section:construction_mask_rc}

In RC, the relation between two entities are represented by a vector. In order to obtain the vector, we confine the attention range of $[CLS]$ token, which is originally used to summarize the overall representation of the sequence, to two entities. Thus, the vector of $[CLS]$ token can accurately summarize the relation between two entities. Based on the dynamic range attention mechanism, we propose two kinds of MASK$^{rc}$ denoted as Eq. \eqref{equation:mask_rc1} and \eqref{equation:mask_rc2}, respectively.
\begin{align}
\label{equation:mask_rc1}
&MASK_{i,j}^{rc}=
  \begin{cases}
     1 &\text{if $i \in P_{CLS}$, $j \in P_{CLS,EN1,EN2}$}\\
     1 &\text{if $i \not\in P_{CLS}$}\\
	 0 &\text{else}
  \end{cases} \\
\label{equation:mask_rc2}
&MASK_{i,j}^{rc}=
  \begin{cases}
     1 &\text{if $i,j \in P_{CLS,EN1,EN2}$}\\
	 0 &\text{else}
  \end{cases}
\end{align}
where $P_{x}$ represents the position of $x$ in sequence S.

The difference between the two matrices is whether the attention range of entity 1 and 2 is confined. In Eq. \eqref{equation:mask_rc1}, the attention range of entity 1 and 2 is not confined, which leads to the vector of RC shifting to the context information of entity. Relatively, in Eq. \eqref{equation:mask_rc2}, only $[CLS]$, entity 1 and 2 are able to pay attention to each other, leading the vector of RC shifting to the information of entity itself. Corresponding to the RC task on medical text, the two MASK matrices will be further analyzed in Section \ref{section:hyperparameter-analysis}.

\subsubsection{The Construction of RC Downstream Task Layer}
\label{section:construction_downstream_rc}

For RC, the downstream task layer needs to convert the representation vector of $[CLS]$ token in the output of STR-encoder into the probability distribution of corresponding relation type. In this paper, we use multilayer perceptron (MLP) to perform this conversion. Specifically, the vector is converted to the probability distribution through two perceptrons with $Tanh$ and $Softmax$ as the activation function, respectively:
\begin{align}
    H_p^{rc}=&Tanh(H_N^{rc}[0]\times W_{rc1}+b_{rc1}) \\
    p_{rc}(R|&S,MASK^{rc},MASK^{all})= \nonumber \\
    &Softmax(H_p^{rc}\times W_{rc2}+b_{rc2})
\end{align}

The training is to minimize loss function $L_{rc}$, denoted as Eq. \eqref{equation:loss_rc}, where $R'$ indicates the real relation type.
\begin{equation}
\label{equation:loss_rc}
\begin{aligned}
    &L_{rc} = -Log[p_{rc}(R'|S,MASK^{rc},MASK^{all})]
\end{aligned}
\end{equation}

\subsection{Joint Learning}

Note that, the parameters are shared in the model except the downstream task layers of NER and RC, which enables STR-encoder to learn the joint features of entities and relations. Moreover, compared with the existing parameter sharing model (e.g., Joint-Bi-LSTM \cite{zheng2017joint}), the feature representation ability of STR-encoder is improved by the feature extraction ability of BERT and its knowledge obtained through pre-training. The loss function of the joint model (i.e., $L_{all}$) will be obtained as follows:
\begin{equation}
    L_{all} = L_{ner} + L_{rc}
\end{equation}
where $L_{ner}$ and $L_{rc}$ are defined in Eq. \eqref{equation:loss_ner} and \eqref{equation:loss_rc}, respectively.

% is defined in Eq. \eqref{equation:loss_ner}, and $L_{rc}$ is defined in Eq. \eqref{equation:loss_rc}.

\section{Experimental Studies}
\label{section:experimental-studies}

\subsection{Dataset and Evaluation Metrics}

The dataset of entity and relation extraction is collected from coronary arteriography reports in Shanghai Shuguang Hospital. There are five types of entities, i.e., Negation, Body Part, Degree, Quantifier and Location. Meanwhile, five relations are included, i.e., Negative, Modifier, Position, Percentage and No Relation. 85\% of ``No Relation" in the dataset are discarded for balance purpose. The statistics of the entities and relations are demonstrated in Table \ref{table:statistics}.

\begin{table}[htbp]
\caption{statistics of different types of entities and relations}
\label{table:statistics}
\begin{center}
\setlength{\tabcolsep}{1.2mm}{
\begin{tabular}{|l|c|l|c|c|}
\hline
Entity Type   & Number       & Relation Type               & Direction          & Number   \\ \hline
Negation      & 103          & Negative                    & e2 to e1                    & 406      \\
Body Part     & 492          & Modifier                    & e2 to e1                    & 1,068    \\
Degree        & 658          & Position                    & e1 to e2                    & 389      \\
Quantifier    & 422          & Percentage                  & bi-direction               & 100 / 256  \\
Location      & 461          & No Relation                 & none                        & 1,975 \\ \hline
Total         & 2,136        & Total                       & none                        & 4,194  \\ \hline
\multicolumn{5}{l}{\begin{tabular}[c]{@{}l@{}}\begin{minipage}{6.8cm}\vspace{1mm}\tiny \item[*] The bi-direction indicates there are two directions, i.e., e1 to e2 and e2 to e1.\end{minipage} \end{tabular}}
\end{tabular}}
\end{center}
\end{table}

In order to ensure the effectiveness of our experiment, we divide the dataset into training, development and test in the ratio of 8:1:1. In the following experiments, we use common performance measures such as Precision, Recall, and F$_1$-score \cite{liu2014strategy} to evaluate NER, RC and joint models.

\subsection{Experimental Setup}

The training of focused attention model proposed in this paper can be divided into two stages. In the first stage, we need to pre-train the shared parameter layer. Due to the high cost of pre-training BERT, we directly adopted parameters pre-trained by Google in Chinese general corpus. In the second stage, we need to fine-tune NER and RC tasks jointly. Parameters of the two downstream task layers are randomly initialized. The two hyperparameters $K$ and MASK$^{rc}$ in the model will be further studied in Section \ref{section:hyperparameter-analysis}.

\subsection{Experimental Result}

To evaluate the performance of our focused attention model, we compare it with state-of-the-art methods on the task of NER, RC and joint entity and relation extraction, respectively.

Based on NER, we experimentally compare our focused attention model with other reference algorithms. These algorithms consist of two NER models in medical domain (i.e., Bi-LSTM \cite{gridach2017character} and RDCNN \cite{qiu2019chinese}) and one joint model in generic domain (i.e., Joint-Bi-LSTM \cite{zheng2017joint}). In addition, we originally plan to use the joint model \cite{li2017neural} in the medical domain, but the character-level representations cannot be implemented in Chinese. Therefore, we replace it with a generic domain model \cite{zheng2017joint} in similar structure. As demonstrated in Table \ref{table:ner-comparison}, the proposed model achieves the best performance, and its precision, recall and F$_1$-score reach 96.69\%, 97.09\% and 96.89\%, which outperforms the second method by 0.2\%, 0.40\% and 1.20\%, respectively.

\begin{table}[!ht]
\caption{Comparisons with the different methods on the task of NER}
\label{table:ner-comparison}
\begin{center}
\setlength{\tabcolsep}{1.2mm}{
\begin{tabular}{|l|c|c|c|}
\hline
\multirow{2}{*}{Methods}             & \multicolumn{3}{c|}{NER}                                  \\ \cline{2-4}
                                     & Precision          & Recall           & F$_1$-score       \\ \hline
Bi-LSTM \cite{Huang2015Bidirectional} & 94.46            & 94.07          & 94.26           \\
RDCNN \cite{qiu2019chinese}           & 96.49            & 94.90          & 95.69           \\
Joint-Bi-LSTM \cite{zheng2017joint}   & 93.84            & 96.69          & 95.24           \\
\textbf{Our model}                  & \textbf{96.69}   & \textbf{97.09} & \textbf{96.89}  \\ \hline
\end{tabular}}
\end{center}
\end{table}

To further investigate the effectiveness of the proposed model on RC, we use two RC models in medical domain (i.e., RCN \cite{wang2018automatic} and CNN \cite{sahu-etal-2016-relation}) and one joint model in generic domain (i.e., Joint-Bi-LSTM \cite{zheng2017joint}) as baseline methods. Since RCN and CNN methods are only applied to RC tasks and cannot extract entities from the text, so we directly use the correct entities in the text to evaluate the RC models. Table \ref{table:rc-comparison} illustrate that the focused attention model achieves the best performance, and its precision, recall and F$_1$-score reach 96.06\%, 96.83\% and 96.44\%, which beats the second model by 1.57\%, 1.59\% and 1.58\%, respectively.

\begin{table}[!ht]
\caption{Comparisons with the different methods on the task of RC}
\label{table:rc-comparison}
\begin{center}
\setlength{\tabcolsep}{1.2mm}{
\begin{tabular}{|l|c|c|c|}
\hline
\multirow{2}{*}{Methods}            & \multicolumn{3}{c|}{RC with Correct Entities}         \\ \cline{2-4}
                                    & Precision          & Recall         & F$_1$-score     \\ \hline
RCN \cite{wang2018automatic}        & 90.77              & 93.65          & 92.19           \\
CNN \cite{nguyen2015relation}       & 94.49              & 95.24          & 94.86           \\
Joint-Bi-LSTM \cite{zheng2017joint} & 92.92              & 92.86          & 92.88           \\
\textbf{Our model}                  & \textbf{96.06}     & \textbf{96.83} & \textbf{96.44}  \\ \hline
\end{tabular}}
\end{center}
\end{table}

For the task of joint entity and relation extraction, we use Joint-Bi-LSTM \cite{zheng2017joint} as baseline method. Since these two models are joint learning, we use the entities predicted in NER as the input for RC. From Table \ref{table:joint-comparison}, we observe that our focused attention model achieves the best performance, and its F$_1$-score reaches 96.89\% and 88.51\%, which is 1.65\% and 1.22\% higher than the second method, respectively. These observations confirm that the feature representation of STR-encoder is indeed stronger than existing common models.

\begin{table}[!ht]
\caption{Comparisons with the state-of-the-art methods on the task of joint entity and relation extraction}
\label{table:joint-comparison}
\begin{center}
\setlength{\tabcolsep}{1.2mm}{
\begin{tabular}{|l|c|c|c|c|c|c|}
\hline
\multirow{2}{*}{Methods}           & \multicolumn{3}{c|}{NER}                                  & \multicolumn{3}{c|}{RC with Predicted Entities}           \\ \cline{2-7}
                                   & Precision          & Recall           & F$_1$-score       & Precision          & Recall           & F$_1$-score       \\ \hline
Joint-Bi-LSTM & 93.84            & 96.69          & 95.24           & 93.64            & 81.75          & 87.29           \\
\textbf{Our model}                & \textbf{96.69}   & \textbf{97.09} & \textbf{96.89}  & \textbf{95.41}   & \textbf{82.54} & \textbf{88.51}  \\ \hline
\end{tabular}}
\end{center}
\end{table}

\section{Experimental Analysis}
\label{section:experimental-analysis}

In this section, we perform additional experiments to analyze the influence of different settings on segmentation points $K$, and different settings on MASK$^{rc}$ and joint learning.

\subsection{Hyperparameter Analysis}
\label{section:hyperparameter-analysis}

We further study the impacts of different settings on segmentation points $K$ defined in Section \ref{section:focused-attention-model} and different settings on MASK$^{rc}$ defined in Section \ref{section:construction_mask_rc}. As shown in Table \ref{table:hyperparameter-analysis}, when $K = 4$ and MASK$^{rc}$ use Eq. \eqref{equation:mask_rc2}, RC reaches the best F$_1$-score of 92.18\%. When $K = 6$ and MASK$^{rc}$ use Eq. \eqref{equation:mask_rc1}, NER achieves the best F$_1$-score of 96.77\%. One possible reason is that MASK$^{rc}$ defined in Eq. \eqref{equation:mask_rc1} doesn't confine the attention range of entity 1 and 2, which enables the model to further learn context information in shared parameter layer, leading to a higher F$_1$-score for NER. For RC, the F$_1$-score with $K = 4$ is the lowest when MASK$^{rc}$ uses Eq. \eqref{equation:mask_rc1}, and reaches the highest when MASK$^{rc}$ uses Eq. \eqref{equation:mask_rc2}. One possible reason is that the two hyperparameters are closely related to each other. However, how they interact with each other in the focus attention model is still an open question.

\begin{table}[!ht]
\caption{Comparisons with different hyperparameters on the task of joint entity and relation extraction}
\label{table:hyperparameter-analysis}
\begin{center}
\setlength{\tabcolsep}{1.2mm}{
\begin{tabular}{|c|c|c|c|c|c|c|c|}
\hline
\multirow{2}{*}{K}          & \multirow{2}{*}{MASK}          & \multicolumn{3}{c|}{NER}                                  & \multicolumn{3}{c|}{RC with Predicted Entities}           \\ \cline{3-8}
                            &                                & Precision          & Recall           & F$_1$-score       & Precision          & Recall           & F$_1$-score       \\ \hline
2                           &Eq. \eqref{equation:mask_rc1}   & 95.07            & 97.93          & 96.48           & 97.08            & \textbf{87.37} & 91.97           \\
4                           &Eq. \eqref{equation:mask_rc1}   & 94.98            & 98.20          & 96.56           & 97.06            & 86.84          & 91.67           \\
6                           &Eq. \eqref{equation:mask_rc1}   & \textbf{95.39}   & \textbf{98.20} & \textbf{96.77}  & 98.20            & 86.32          & 91.88           \\
2                           &Eq. \eqref{equation:mask_rc2}   & 94.67            & 97.93          & 96.27           & 95.95            & \textbf{87.37} & 91.46           \\
4                           &Eq. \eqref{equation:mask_rc2}   & 94.77            & 97.87          & 96.29           & \textbf{98.21}   & 86.84          & \textbf{92.18}  \\
6                           &Eq. \eqref{equation:mask_rc2}   & 94.47            & 97.93          & 96.17           & 96.51            & \textbf{87.37} & 91.71           \\ \hline
\end{tabular}}
\end{center}
\end{table}

\subsection{Ablation Analysis}

In order to evaluate the influence of joint learning, we train NER and RC models separately as an ablation experiment. In addition, we also use correct entities to evaluate RC, excluding the effect of NER results on the RC results, and independently compare the NRE and RC tasks.

\begin{table}[!ht]
\setlength{\abovecaptionskip}{0pt}
\setlength{\belowcaptionskip}{0pt}
\caption{Comparisons with training NER and RC tasks separately}
\label{table:ablation-analysis}
\begin{center}
\setlength{\tabcolsep}{1.2mm}{
\begin{tabular}{|l|c|c|c|c|c|c|}
\hline
\multirow{2}{*}{Methods}          & \multicolumn{3}{c|}{NER}                                  & \multicolumn{3}{c|}{RC with Correct Entities}             \\ \cline{2-7}
                                  & Precision          & Recall           & F$_1$-score       & Precision          & Recall           & F$_1$-score       \\ \hline
Only NER                          & 95.14            & \textbf{97.64} & 96.37           & -                  & -                & -                 \\
Only RC                           & -                  & -                & -                 & 96.00            & 95.24          & 95.62           \\
\textbf{Our model}               & \textbf{96.69}   & 97.09          & \textbf{96.89}  & \textbf{96.06}   & \textbf{96.83} & \textbf{96.44}  \\ \hline
\end{tabular}}
\end{center}
\end{table}

As shown in Table \ref{table:ablation-analysis}, compared with training separately, the results are improved by 0.52\% score in F$_1$-score for NER and 0.82\% score in F$_1$-score for RC. It shows that joint learning can help to learn the joint features between NER and RC and improves the accuracy of two tasks at the same time. For NER, precision score is improved by 1.55\%, but recall score is reduced by 0.55\%. One possible reason is that, although the relationship type can guide the model to learn more accurate entity types, it also introduces some uncontrollable noise. In summary, joint learning is an effective method to obtain the best performance.

\section{Conclusion}
\label{section:conclusion-future-work}

In order to structure medical text, entity and relation extraction is an indispensable step. In this paper, we propose a focused attention model to jointly learn NER and RC task based on a shared task representation encoder which is transformed from BERT through dynamic range attention mechanism. Compared with existing models, our model can extract the entities and relations from the medical text more accurately. The experimental results on coronary angiography texts verify the effectiveness of our model.

\section{Acknowledgment}

The authors would like to appreciate the efforts of the editors and valuable comments from the anonymous reviewers. This work is supported by the National Key R\&D Program of China for ``Precision Medical Research" (Grant No. 2018YFC0910500), the National Natural Science Foundation of China (Grant No. 61772201), the Special Fund Project for ``Shanghai Informatization Development in Big Data" (Grant No. 201901043) and the Open Fund of Shanghai Key Laboratory of Multidimensional Information Processing, East China Normal University (Grant No. 2019MIP004).

\bibliographystyle{IEEEtran}
\bibliography{Bibfiles}

\end{document}